\documentclass[runningheads]{llncs}
\usepackage{graphicx}
\usepackage{multirow}
\usepackage{array}
\usepackage{float}
\usepackage{hyperref}
\usepackage{colortbl}
\begin{document}
\title{Self-Prompting Large Vision Models for Few-Shot Medical Image Segmentation}
\titlerunning{Self-Prompting Large Vision Models}
%
\author{Qi Wu\inst{\star,1} \and Yuyao Zhang\inst{\star,1} \and Marawan Elbatel\inst{1,2} 
}
\renewcommand{\thefootnote}{\fnsymbol{footnote}}
\footnotetext[1]{Co-first authors}

\authorrunning{Q. Wu et al.}


\institute{The Hong Kong University of Science and Technology \and Computer Vision and Robotics Institute, University of Girona \\
\email{qwuaz@connect.ust.hk, yzhangkp@connect.ust.hk}}

\maketitle              
\begin{abstract}
Recent advancements in large foundation models have shown promising potential in the medical industry due to their flexible prompting capability. One such model, the Segment Anything Model (SAM), a prompt-driven segmentation model, has shown remarkable performance improvements, surpassing state-of-the-art approaches in medical image segmentation. However, existing methods primarily rely on tuning strategies that require extensive data or prior prompts tailored to the specific task, making it particularly challenging when only a limited number of data samples are available. In this paper, we propose a novel perspective on self-prompting in medical vision applications. Specifically, we harness the embedding space of SAM to prompt itself through a simple yet effective linear pixel-wise classifier. By preserving the encoding capabilities of the large model, the contextual information from its decoder, and leveraging its interactive promptability, we achieve competitive results on multiple datasets (i.e. improvement of more than 15\% compared to fine-tuning the mask decoder using a few images). Our code is available at~\href{https://github.com/PeterYYZhang/few-shot-self-prompt-SAM}{https://github.com/PeterYYZhang/few-shot-self-prompt-SAM}

\end{abstract}
\section{Introduction}
Supervised methods in medical image analysis require significant amounts of labeled data for training, which can be costly and impractical due to the scarcity of high-quality labeled medical data. To solve this problem, many works like ~\cite{cai2020few,singh2021metamed}, adopted
few-shot learning, which aims at generalizing model to a new class via learning from a small number of samples. However, such methods tried to learn contextual information of the new class, which is hard since the contextual information can be complex and multi-faceted, and can be easily influenced by noise. Therefore, we seek for methods that require less information, such as the size and location of the segmentation target.

Recent advancements in large-scale models, such as GPT-4~\cite{openai2023gpt4}, DALL-E~\cite{ramesh2021zeroshot}, and SAM~\cite{kirillov2023segment}, have shed light on few-shot and even zero-shot learning. Due to their remarkable capabilities in transferring to multiple downstream tasks with limited training data, these models can act as foundation models with exceptional generalization abilities, and prompts play a crucial role in determining their overall performances. Many downstream tasks benefit from these models by leveraging prompt engineering~\cite{zhou2023can,Elbatel2023FoProKDFP} and fine-tuning techniques~\cite{ma2023segment,zhang2023customized,wu2023medical}

One prominent large foundation model in computer vision, the Segment Anything Model (SAM)~\cite{kirillov2023segment}, is a powerful tool for various segmentation tasks, trained on natural images. The model can generate different masks based on different user input prompts.
Due to SAM's promptable nature, it can potentially assist medical professionals in interactive segmentation tasks. 

When solving practical tasks in clinics and hospitals, several challenges need to be addressed: 1) How to tackle the scarcity of medical data, and 2) how to be user-friendly and assist medical professionals in more flexible way.

However, though typical few-shot learning models can reduce the data required, they do not have the promptable feature. Also, while some other SAM fine-tuning method~\cite{ma2023segment,zhang2023customized} can achieve promptability, more labelled data are required during training.

Recently, self-prompting arises in tuning large language models  (LLMs)~\cite{li2022self}, where the model prompts itself to improve the performance. 
To overcome the two aforementioned challenges simultaneously, we draw inspiration from the success of self-prompting LLMs and propose a novel method that utilizes a simple linear pixel-wise classifier to self-prompt the SAM model. 
Our method leverage the promptable feature of large vision foundation model, having a simple architecture by inserting a small plug-and-play unit in SAM. At the same time, all the training can be done with limited labelled data and time. 
Remarkably, our method can already achieve good results using only a few images training set, outperforming some other fine-tuning methods~\cite{ma2023segment,zhang2023customized} that use the same amount of data. Furthermore, our method is almost training-free, the training can be done within 30 seconds while other fine-tuning methods require more than 30 minutes (in the few-shot setting). This allows generation of output masks with few computational resources and time, which can assist medical professionals in generating more precise prompts or labeling data.

To summarize, our major contributions are: 

\begin{itemize}
  \item We propose a novel computational efficient method that leverages the large-scaled pre-trained model SAM for few-shot medical image segmentation
  \item We develop a method to self-prompt the foundation model SAM in the few-shot setting and demonstrated the potential and feasibility of such self-prompting method for medical image segmentation
  \item We experiment and show that our method outperforms other SAM fine-tuning methods in a few-shot setting and is more practical in clinical use
\end{itemize}

\section{Related Works}
\subsection{Few-shot Medical Image Segmentation}
Few-shot learning has been popular in medical image segmentation, as it requires significantly less data while still reaching satisfactory results. Previous works~\cite{wang2022few,makarevich2021metamedseg,sun2022few,cai2020few,feyjie2020semi,singh2021metamed} have shown great capacity for few-shot learning in some different medical segmentation tasks. But this methods aimed to learn the prototype knowledge or contextual information of the target domain, which is easily influenced by noise and other factors since those information are complex and multi-faceted. Unlike these methods, we propose a novel technique that utilizing a large vision foundation model to achieve few-shot learning. 
\subsection{SAM}
Inspired by the "prompting" techniques of NLP's foundation models~\cite{devlin2018bert}, the project team defined the segment anything task as returning valid segmentation masks given any segmentation prompts. To introduce zero-shot generalization in segmentation, the team proposed the Segment Anything Model (SAM)~\cite{kirillov2023segment}, which is a large-scaled vision foundation model. 
Several recent studies, including~\cite{deng2023segment,ji2023segment,zhou2023can,hu2023sam,mohapatra2023brain,he2023accuracy,mattjie2023exploring}, have evaluated SAM's capability on different medical image segmentation tasks in the context of zero-shot transfer. The results show that SAM can generate satisfactory masks with sufficient high-quality prompts for certain datasets. However, manually and accurately prompting SAM from scratch will be time-consuming and inefficient for medical professionals. Our method can self-generate prompts, hence being more user-friendly and can assist professionals during inference.

\subsection{Tuning The Segment Anything Model}
Large Foundation models in vision including SAM are difficult to be trained or tuned from scratch due to the limitation of computing resources and data. Multiple previous works~\cite{ma2023segment,zhang2023customized,wu2023medical} have fine-tuned SAM on medical datasets. Specifically, MedSAM~\cite{ma2023segment} fine-tune the SAM mask decoder on a large-scaled datasets, SAMed~\cite{zhang2023customized} adopt low-rank-based fine-tuning strategy (LoRA)~\cite{hu2021lora} and train a default prompt for all image in the dataset, Medical SAM Adapter (MSA)~\cite{wu2023medical} use adapter modules for fine-tuning. These methods yield satisfactory results, getting close to or even outperforming SOTA fully-supervised models.
However, these SAM-based works still needs large amounts of data to fine-tune the model in a supervised way, yet have not fully leverage the prompting ability. 

\section{Methodology}
We denote the training dataset as $D=\displaystyle{\{ I, T \}}$, where $I=\{i_1,\dots,i_n\}$, and $T=\{t_1,\dots,t_n\}, \ n \in N$  corresponding to the images and segmentation ground truth. Our goal is to design an plug-and-play self-prompting unit that can provide SAM with the \textbf{location} and \textbf{size} information of the segmentation target with only a \textbf{few labeled data}, $k$ images for example, denoted as $D_k=\{I_k, T_k\}$. 
\begin{figure}[h]
    \centering
    \includegraphics[width=\textwidth]{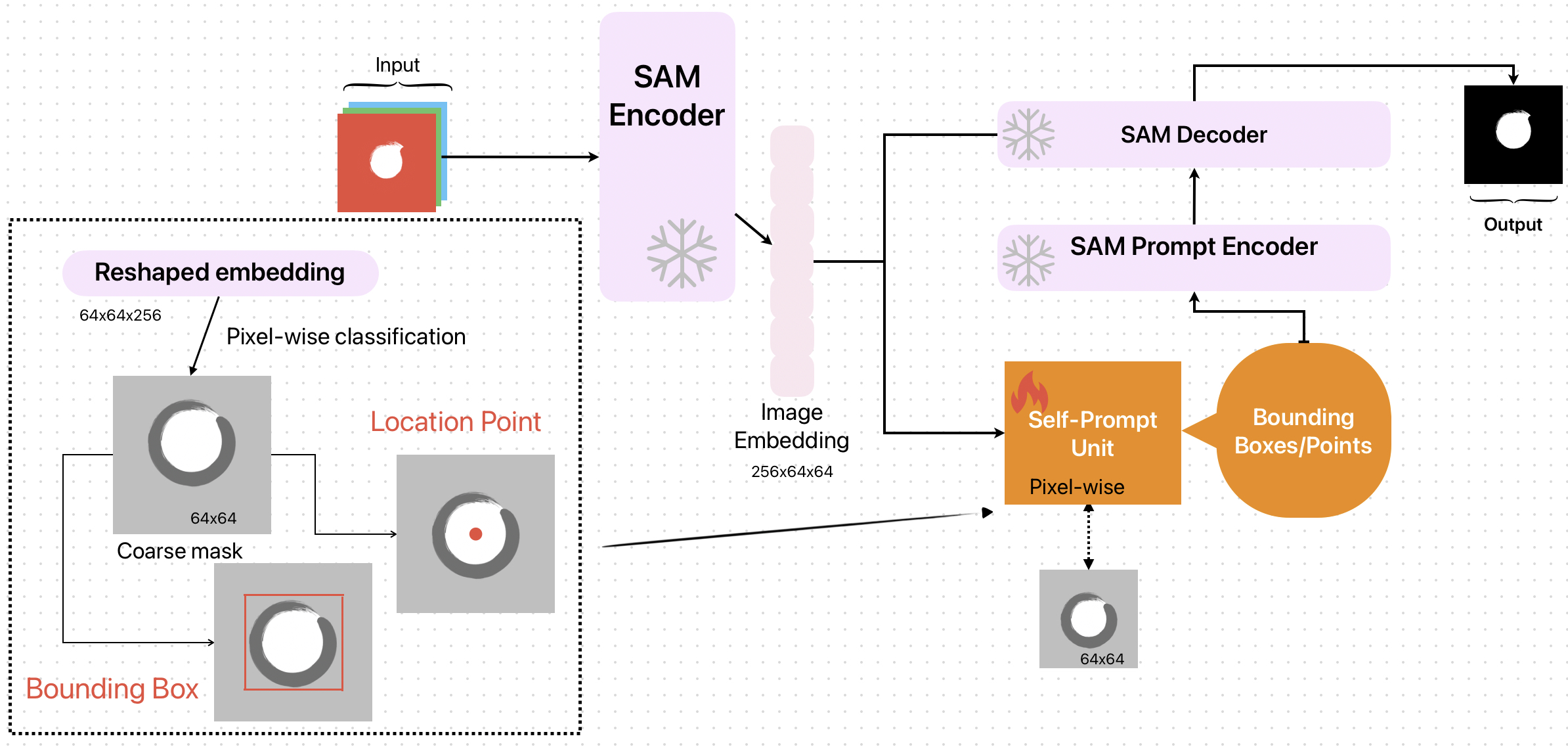}
    \caption{The overall design of our framework. The pink modules are exactly the same as the original SAM architecture, and the "snowflake" sign represents that we freeze the module during training. The Self-Prompt Unit ("fire" sign) is trained using the image embeddings from the SAM encoder and the resized ground truth label. The unit predicts a coarse mask, which is used to obtain the bounding box and location point that prompt SAM.} 
    \label{fig:framework}
\end{figure}
As shown in Fig.~\ref{fig:framework}, our model build upon the original SAM model(the pink blocks) which is kept frozen all the time during training and inference. For each image $i_n$, the image encoder maps it to the embedding spaces $z_n=E_{SAM}(i_n)$, then our self-prompt unit take the image embedding to provide the bounding box and point as prompt $p_n$. Finally, the decoder combined the encoded prompt $E_{prompt}(p_n)$ and the image embedding $z_n$ to get the final segmentation output.

\subsection{Self-Prompt Unit}
To learn the location and size information of the target, an intuitive way is to get a coarse mask as a reference. After passing through the powerful encoder of SAM, which is a Vision Transformer (ViT)~\cite{dosovitskiy2020image}, the input image is encoded as a vector $z_n\in R^{256\times64\times64}$. To align the mask with the encoded image embedding, we down-sample it to $64\times64$. Here the mask is treated as binary, then we conduct a logistic regression to classify each pixel as background or mask to get the coarse mask. Also, using a logistic regression instead of neural networks will minimize the influence of inference speed. Finally, from the predicted low-resolution mask, the location point and the bounding box can be obtained using morphology and image processing techniques.
\subsubsection{Location Point}
We use distance transform to find one point inside the predicted mask to represent the location. Distance transform is an image processing technique used to compute the distance of each pixel in an image to its nearest boundary. After distance transform, the value of each pixel is replaced with its Euclidean distance to the nearest boundary. We can obtain the point that is farthest from the boundary by finding the pixel with the maximum distance.
\subsubsection{Bounding Boxes}
The bounding boxes are generated using the minimum and maximum X, Y coordinates of the predicted mask generated by the linear pixel-wise classifier, and added by a 0-20 pixels' perturbation. Due to the simpleness of the linear layer, the original outputs are not high-quality. Noises and holes occurs in the masks. To overcome this, we add some simple morphology processes, erosion and dilation, on the outputs of the linear classifier. And the refined masks are used for prompts generation.

\subsection{Training Objectives}
Each image-mask pair in the training dataset $D_k=\{I_k, T_k\}$ is denoted as $\{i^q,t^q\}$, for simplicity we denote the down-sampled mask the same as the original one. The image $i^q$ is first fed to the image encoder to get the image embedding $z^q\in R^{256\times64\times64}$ then is reshaped to $z^q\in R^{64\times64\times256}$ in corresponding to the mask of shape $t^q\in R^{64\times64}$. Since we perform the Logistic Regression pixel-wisely, the loss function becomes,
\begin{equation}
    L=\frac{1}{k}\sum_q^k\sum_{1\leq m, n\leq 64}-(t_{m,n}^q \log\hat{t}_{m,n}^q+(1-t_{m,n}^q\log(1-\hat{t}_{m,n}^q)))
\end{equation}
where, $t_{m,n}^q$ is the value of pixel $(m,n)$ of the $q$-th mask in the subset $D_k$, and $\hat{t}_{m,n}^q = \mathbf{1}_{\sigma(w^Tz^q_{m,n}+b)>0.5}$, where $z^q_{m,n}, w, b \in R^{256}$, and $\sigma$ is the Sigmoid function.

\section{Experiments}
\subsection{Datasets}
We split the datasets into 5 parts with equal size and apply k-fold (k=5) cross-validation to evaluate our method. 

\noindent\textbf{{Kvasir-SEG}}
The Kvasir-SEG dataset~\cite{jha2020kvasir} contains 1000 polyp images and their corresponding ground truth. It has JPEG encoding and includes bounding box coordinates in a JSON file. The task on this dataset is to segment polyps. 

\noindent\textbf{ISIC-2018}
The ISIC-2018 dataset~\cite{tschandl2018data,codella2018skin} was published by the International Skin Imaging Collaboration (ISIC) as a comprehensive collection of dermoscopy images. It is an aggregate dataset that comprises a total of 2594 images, providing a diverse range of examples for analysis. This dataset is for skin lesion segmentation.

\subsection{Implementation Details}
We use the ViT-B SAM model, the smallest version of SAM, as our backbone.

First, we use the SAM image encoder (ViT-B) to obtain the image embeddings. In the embedding space, the spatial resolution is 64x64 pixels and each pixel corresponds to a 256-dimensional vector that represents the features extracted from the original image. To train a pixel-wise classifier, we utilize these pixel-wise embeddings alongside the ground truth labels. The ground truth labels are resized to 64x64 to match the resolution of the embeddings. Each pixel in this resized label image corresponds to a specific class.

For simplicity, we utilize the logistic regression module directly from the scikit-learn~\cite{scikit-learn} library. We set the maximum iteration to 1000 and use the default values for all other hyper-parameters. We make this decision because we are only interested in generating simple prompts and believe that intricate hyper-parameter tuning would be unnecessary and will not significantly improve our results. By doing this, we can focus on creating and testing our prompts without tuning the hyper-parameters for different datasets.

During the inference phase, the linear pixel-wise classifier takes an image embedding as input and produces a low-resolution mask of size 64x64. This mask is then resized to 256x256 and subjected to a 3-iteration erosion followed by a 5-iteration dilation using a 5x5 kernel. The resulting mask is used to generate prompts, including a point and a bounding box, for further processing. The prompts, together with the image embeddings, are then passed to the SAM mask decoder to generate the final mask.

\subsection{Results}
We use Dice and IoU as our evaluation metrics. We compare our method with another two fine-tuning models, MedSAM~\cite{ma2023segment} and SAMed~\cite{zhang2023customized}. We also include the result of the original unprompted SAM-B by extracting the masks that best overlap the evaluated ground truth (same as the setting in~\cite{zhou2023can}). Notice that we did not utilize the ground truth in any mask extraction for evaluating our method. For MedSAM, we set the bounding box (xmin, ymin, xmax, ymax) as the image size (0, 0, H, W) for all images, and trained the decoder. For SAMed, we only fine-tuned the decoder for fairness. We set the size of the training set to be 20. The reason for 20 shots instead of fewer shots is that, it is hard for the other two fine-tuning methods to generate valid masks with such limited training data. We want to obtain a more valuable comparison. The performance of our approach using less number of shots remains competitive, which we demonstrate in the ablation study.

The results are shown in table~\ref{tab1} and some examples are in fig.~\ref{fig:examples_kavir}. Our method using both the point and bounding box reaches 62.78\% and 53.36\% in Dice and IoU score on Kvasir-SEG and 66.78\% and 55.32\% on ISIC2018 correspondingly, which surpasses MedSAM~\cite{ma2023segment}, SAMed~\cite{zhang2023customized} and the original unprompted SAM. Notice that training full data on ISIC2018, the metrics are a lot higher than our methods trained on 20 images, this may because the dataset is quite large, more than 2000 images. More discussion of our results will be presented in ablation study.

Remarkably, the training of our method can be done in 30 seconds using an NVIDIA RTX-3090 GPU. In fact, most of the time is used for computing image embeddings. If the embeddings are pre-computed, the training can be done within a few seconds using a CPU. In comparison, for other fine-tuning methods, at least 30 minutes are required in the few-shot setting. Some limitations of our work are discussed in the supplementary.

\begin{table}
\caption{Comparison of our method (using both of the bounding box and the point) to other fine-tuning methods in a few-shot setting. Number of shots is set to be 20 here. We use the the ground truth to extract the best masks for unprompted SAM, the prompted one is an upper bound in which prompts are generated from the ground truth. The results of other methods fine-tuned on the whole dataset are also listed for reference. Notice that, for SAMed, we only trained the decoder. The U-net here is pre-trained on ImageNet.
}\label{tab1}
\begin{center}
\begin{tabular}{|c|>{\centering\arraybackslash}m{2cm}|>{\centering\arraybackslash}m{2cm}|>{\centering\arraybackslash}m{2cm}|>{\centering\arraybackslash}m{2cm}|}
\hline
\multirow{2}{*}{Models} & \multicolumn{2}{|c|}{Kvasir-SEG} & \multicolumn{2}{|c|}{ISIC2018} \\
\cline{2-5}
&Dice\%$\uparrow$ &IoU\%$\uparrow$ &Dice\%$\uparrow$ &IoU\%$\uparrow $\\
\hline
Ours & \textbf{62.78} & \textbf{53.36} & \textbf{66.78} & \textbf{55.32} \\
\hline
MedSAM~\cite{ma2023segment} & 55.01 & 43.21 & 64.94 & 54.65 \\
\hline
SAMed~\cite{zhang2023customized} & 61.48 & 51.75 & 63.27 & 54.23 \\
\hline
Unprompted SAM-B~\cite{zhou2023can} & 52.66 & 44.27 & 45.25 & 36.43 \\
\arrayrulecolor{red}\hline\arrayrulecolor{black}
MedSAM~\cite{ma2023segment} (full-data) & 66.14 & 55.77 & 84.22 & 75.49 \\
\hline
SAMed~\cite{zhang2023customized} (full-data) & 79.99 & 71.07 & 85.49 & 77.40 \\
\hline
Prompted SAM-B & 86.86 & 79.49 & 84.28 & 73.93 \\
\hline
U-net~\cite{ronneberger2015u} (full-data) & 88.10 & 81.43 & 88.36 & 81.14 \\
\hline
\end{tabular}
\end{center}
\end{table}

\section{Ablation Study}
\subsection{Number of shots}
We perform ablation study on the number of shots ($k$) of our model. We choose $k=10,20,40$ and full dataset. The outcomes obtained on ISIC-2018 are presented in Table~\ref{tab2}, whereas the results of Kvasir-SEG are provided in the supplementary materials. The overall performance is improved with more shots. However, the improvement is limited. This is understandable since we only train a simple logistic regression. Due to the limited performance of the classifier, generating more valid prompts is hard. But if we train a more sophisticated classifier, it will like an decoder and deviate from our main goal of reducing data and computation requirement. 
Iterative use of masks and more advanced prompt generating mechanism can be explored in the future. 

\begin{table}[h]
\caption{Results of the ablation study showing the impact of number of shots and different prompting methods. "Linear" means the coarse mask from the linear pixel-wise classifier; "point" means only using the point as prompt; "box" represents only using the bounding box as prompt;"point+box" means using both of bounding box and point as prompt. The scores here are dice scores.}\label{tab2}
\begin{center}
\begin{tabular}{|c|>{\centering\arraybackslash}m{2cm}|>{\centering\arraybackslash}m{2cm}|>{\centering\arraybackslash}m{2cm}|>{\centering\arraybackslash}m{2cm}|}
\hline
\multicolumn{5}{|c|}{ISIC-2018} \\
\hline
 & 10 shots & 20 shots & 40 shots & full-data\\
\hline
Ours(linear) & 58.81 & 62.69 & 63.81 & 65.62 \\
\hline
Ours(point) & 59.49 & 61.23 & 61.52 & 61.76 \\
\hline
Ours(box) &47.13 & 47.99 & 49.81 & 55.03 \\
\hline
Ours(point+box) & 64.22 & 66.78 & 67.88 & 69.51 \\
\hline
\end{tabular}
\end{center}
\end{table}

\begin{figure}[t]
    \centering
    \includegraphics[width=\textwidth]{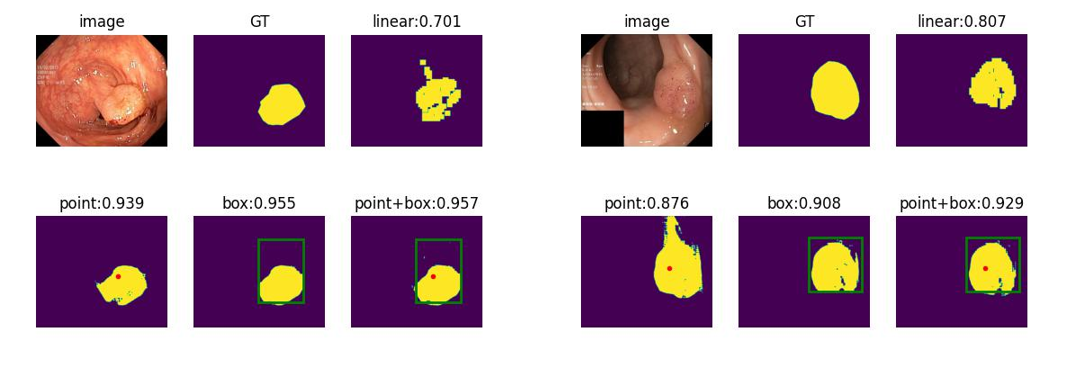}
    \caption{Some examples of using different methods on Kvasir-SEG dataset. The yellow objects in the figures denote the segmented polyps. "Linear" here is the coarse mask from the linear pixel-wise classifier. The scores here are Dice scores.}
    \label{fig:examples_kavir}
\end{figure}

\subsection{Use of methods}
We also conduct the ablation study of the methods: 1) using only the point; 2) using only the bounding box; 3) using both the point and the bounding box. Results are shown in Table~\ref{tab2}.
The bounding box provides the size information and a rough position of the instance, while the point gives the accurate position of the instance. Fig.~\ref{fig:examples_kavir} demonstrates some examples of utilizing different prompt generation approaches. Since we get the coarse mask from the linear pixel-wise classifier, the size information of the instance may be highly influenced by the mistakenly classified pixels (i.e. the left example), in this case, points will help to locate the object. But the point itself cannot provide size information, which may cause the model to generate masks with wrong size (i.e. the right example), in this case boxes will help to restrict the size. Taking the advantages of both the point and the bounding box, we can obtain better results. More results, including failure cases, are included in our GitHub repository.

\section{Conclusion}
Our study has demonstrated the potential and feasibility of utilizing self-prompting with large-scale vision foundation models for medical image segmentation. We present a simple yet effective idea, using a few images to train a linear pixel-wise classifier on the image embedding space to generate a prompt for SAM. Our method can be more user-friendly than traditional few-shot learning models due to SAM's promptable feature and requires significantly less data than those SAM fine-tuning models.
We evaluate our method on two datasets. The results show that our method outperforms another two fine-tuning methods using same amount of data. Since the whole process requires few computational resources and time, the resulting outputs can also be regarded as raw masks that can assist medical professionals in more precise prompt generation and data labeling. Future studies can lay more emphasis on getting more accurate prompts from the output of SAM. Combining self-prompting with other fine-tuning methods can also be explored. 

\vspace{2mm}
\noindent \textbf{Acknowledgement} M.E is partially funded by the EACEA Erasmus Mundus grant. We would like to acknowledge Prof. Xiaomeng Li for revising our manuscript. We would also like to thank Mr. Haonan Wang for providing valuable suggestions to our work. 
%
%
%
\bibliographystyle{splncs04}
\bibliography{references}

\begin{thebibliography}{10}
\providecommand{\url}[1]{\texttt{#1}}
\providecommand{\urlprefix}{URL }
\providecommand{\doi}[1]{https://doi.org/#1}

\bibitem{cai2020few}
Cai, A., Hu, W., Zheng, J.: Few-shot learning for medical image classification.
  In: International Conference on Artificial Neural Networks. pp. 441--452.
  Springer (2020)

\bibitem{codella2018skin}
Codella, N.C., Gutman, D., Celebi, M.E., Helba, B., Marchetti, M.A., Dusza,
  S.W., Kalloo, A., Liopyris, K., Mishra, N., Kittler, H., et~al.: Skin lesion
  analysis toward melanoma detection: A challenge at the 2017 international
  symposium on biomedical imaging (isbi), hosted by the international skin
  imaging collaboration (isic). In: 2018 IEEE 15th international symposium on
  biomedical imaging (ISBI 2018). pp. 168--172. IEEE (2018)

\bibitem{deng2023segment}
Deng, R., Cui, C., Liu, Q., Yao, T., Remedios, L.W., Bao, S., Landman, B.A.,
  Wheless, L.E., Coburn, L.A., Wilson, K.T., et~al.: Segment anything model
  (sam) for digital pathology: Assess zero-shot segmentation on whole slide
  imaging. arXiv preprint arXiv:2304.04155  (2023)

\bibitem{devlin2018bert}
Devlin, J., Chang, M.W., Lee, K., Toutanova, K.: Bert: Pre-training of deep
  bidirectional transformers for language understanding. arXiv preprint
  arXiv:1810.04805  (2018)

\bibitem{dosovitskiy2020image}
Dosovitskiy, A., Beyer, L., Kolesnikov, A., Weissenborn, D., Zhai, X.,
  Unterthiner, T., Dehghani, M., Minderer, M., Heigold, G., Gelly, S., et~al.:
  An image is worth 16x16 words: Transformers for image recognition at scale.
  arXiv preprint arXiv:2010.11929  (2020)

\bibitem{Elbatel2023FoProKDFP}
Elbatel, M., Martí, R., Li, X.: Fopro-kd: Fourier prompted effective knowledge
  distillation for long-tailed medical image recognition. ArXiv
  \textbf{abs/2305.17421} (2023)

\bibitem{feyjie2020semi}
Feyjie, A.R., Azad, R., Pedersoli, M., Kauffman, C., Ayed, I.B., Dolz, J.:
  Semi-supervised few-shot learning for medical image segmentation. arXiv
  preprint arXiv:2003.08462  (2020)

\bibitem{he2023accuracy}
He, S., Bao, R., Li, J., Grant, P.E., Ou, Y.: Accuracy of segment-anything
  model (sam) in medical image segmentation tasks. arXiv preprint
  arXiv:2304.09324  (2023)

\bibitem{hu2023sam}
Hu, C., Li, X.: When sam meets medical images: An investigation of segment
  anything model (sam) on multi-phase liver tumor segmentation. arXiv preprint
  arXiv:2304.08506  (2023)

\bibitem{hu2021lora}
Hu, E.J., Shen, Y., Wallis, P., Allen-Zhu, Z., Li, Y., Wang, S., Wang, L.,
  Chen, W.: Lora: Low-rank adaptation of large language models. arXiv preprint
  arXiv:2106.09685  (2021)

\bibitem{jha2020kvasir}
Jha, D., Smedsrud, P.H., Riegler, M.A., Halvorsen, P., de~Lange, T., Johansen,
  D., Johansen, H.D.: Kvasir-seg: A segmented polyp dataset. In: International
  Conference on Multimedia Modeling. pp. 451--462. Springer (2020)

\bibitem{ji2023segment}
Ji, W., Li, J., Bi, Q., Li, W., Cheng, L.: Segment anything is not always
  perfect: An investigation of sam on different real-world applications. arXiv
  preprint arXiv:2304.05750  (2023)

\bibitem{jieyun_bai_2023_7851339}
Jieyun, B.: {Pubic Symphysis-Fetal Head Segmentation and Angle of Progression}
  (Apr 2023). \doi{10.5281/zenodo.7851339},
  \url{https://doi.org/10.5281/zenodo.7851339}

\bibitem{kirillov2023segment}
Kirillov, A., Mintun, E., Ravi, N., Mao, H., Rolland, C., Gustafson, L., Xiao,
  T., Whitehead, S., Berg, A.C., Lo, W.Y., et~al.: Segment anything. arXiv
  preprint arXiv:2304.02643  (2023)

\bibitem{li2022self}
Li, J., Zhang, Z., Zhao, H.: Self-prompting large language models for
  open-domain qa. arXiv preprint arXiv:2212.08635  (2022)

\bibitem{ma2023segment}
Ma, J., Wang, B.: Segment anything in medical images. arXiv preprint
  arXiv:2304.12306  (2023)

\bibitem{makarevich2021metamedseg}
Makarevich, A., Farshad, A., Belagiannis, V., Navab, N.: Metamedseg: volumetric
  meta-learning for few-shot organ segmentation. arXiv preprint
  arXiv:2109.09734  (2021)

\bibitem{mattjie2023exploring}
Mattjie, C., de~Moura, L.V., Ravazio, R.C., Kupssinsk{\"u}, L.S., Parraga, O.,
  Delucis, M.M., Barros, R.C.: Exploring the zero-shot capabilities of the
  segment anything model (sam) in 2d medical imaging: A comprehensive
  evaluation and practical guideline. arXiv preprint arXiv:2305.00109  (2023)

\bibitem{mohapatra2023brain}
Mohapatra, S., Gosai, A., Schlaug, G.: Brain extraction comparing segment
  anything model (sam) and fsl brain extraction tool. arXiv preprint
  arXiv:2304.04738  (2023)

\bibitem{openai2023gpt4}
OpenAI: Gpt-4 technical report (2023)

\bibitem{scikit-learn}
Pedregosa, F., Varoquaux, G., Gramfort, A., Michel, V., Thirion, B., Grisel,
  O., Blondel, M., Prettenhofer, P., Weiss, R., Dubourg, V., Vanderplas, J.,
  Passos, A., Cournapeau, D., Brucher, M., Perrot, M., Duchesnay, E.:
  Scikit-learn: Machine learning in {P}ython. Journal of Machine Learning
  Research  \textbf{12},  2825--2830 (2011)

\bibitem{ramesh2021zeroshot}
Ramesh, A., Pavlov, M., Goh, G., Gray, S., Voss, C., Radford, A., Chen, M.,
  Sutskever, I.: Zero-shot text-to-image generation (2021)

\bibitem{ronneberger2015u}
Ronneberger, O., Fischer, P., Brox, T.: U-net: Convolutional networks for
  biomedical image segmentation. In: Medical Image Computing and
  Computer-Assisted Intervention--MICCAI 2015: 18th International Conference,
  Munich, Germany, October 5-9, 2015, Proceedings, Part III 18. pp. 234--241.
  Springer (2015)

\bibitem{singh2021metamed}
Singh, R., Bharti, V., Purohit, V., Kumar, A., Singh, A.K., Singh, S.K.:
  Metamed: Few-shot medical image classification using gradient-based
  meta-learning. Pattern Recognition  \textbf{120},  108111 (2021)

\bibitem{sun2022few}
Sun, L., Li, C., Ding, X., Huang, Y., Chen, Z., Wang, G., Yu, Y., Paisley, J.:
  Few-shot medical image segmentation using a global correlation network with
  discriminative embedding. Computers in biology and medicine  \textbf{140},
  105067 (2022)

\bibitem{tschandl2018data}
Tschandl, P., Rosendahl, C., Kittler, H.: Data descriptor: the ham10000
  dataset, a large collection of multi-source dermatoscopic images of common
  pigmented skin lesions. Sci. Data  \textbf{5}(1) (2018)

\bibitem{wang2022few}
Wang, R., Zhou, Q., Zheng, G.: Few-shot medical image segmentation regularized
  with self-reference and contrastive learning. In: International Conference on
  Medical Image Computing and Computer-Assisted Intervention. pp. 514--523.
  Springer (2022)

\bibitem{wu2023medical}
Wu, J., Fu, R., Fang, H., Liu, Y., Wang, Z., Xu, Y., Jin, Y., Arbel, T.:
  Medical sam adapter: Adapting segment anything model for medical image
  segmentation. arXiv preprint arXiv:2304.12620  (2023)

\bibitem{zhang2023customized}
Zhang, K., Liu, D.: Customized segment anything model for medical image
  segmentation. arXiv preprint arXiv:2304.13785  (2023)

\bibitem{zhou2023can}
Zhou, T., Zhang, Y., Zhou, Y., Wu, Y., Gong, C.: Can sam segment polyps? arXiv
  preprint arXiv:2304.07583  (2023)

\end{thebibliography}

\newpage
\appendix
\section{Supplementary Materials}

\subsection{Limitation}
\subsubsection{Multi-instance Segmentation}

The first limitation of our method lies in the segmentation task that has multiple instances. One can blame the problem on the plugged-in linear classifier, the simple classifier cannot know the number of instances accurately, so the spatial information passed to the prompt encoder is not complete, thus leading to the limited performance. More advanced training and prompting techniques need to be explored in the future.

\subsubsection{Limitation of Modality Knowledge in Decoder}

\begin{figure}[h]
    \centering
    \includegraphics[width = 0.7\textwidth]{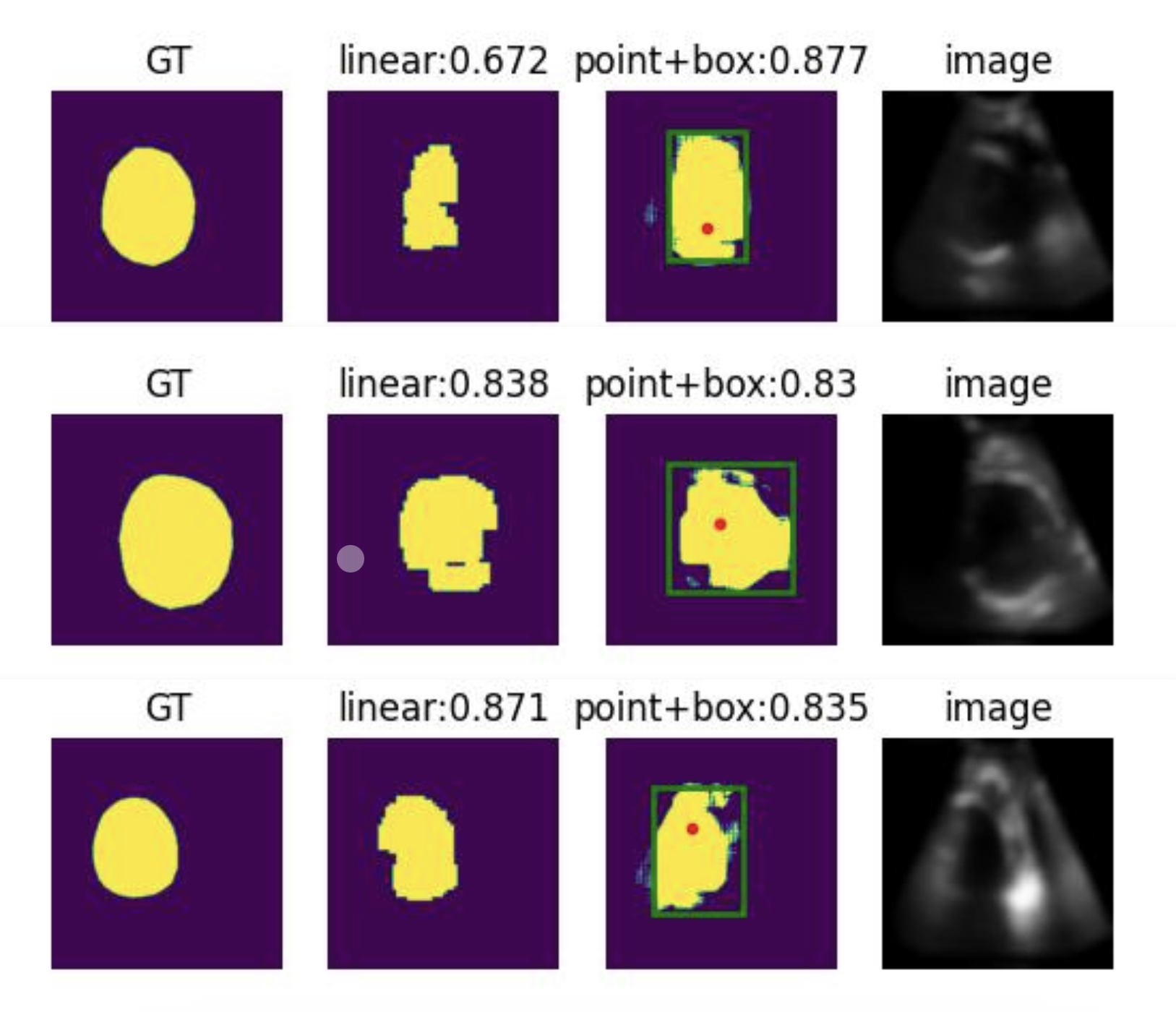}
    \caption{Some examples of the result of our method on the Pubic Symphysis-Fetel dataset. The segmentation result is not satisfactory in ultrasound images, although the score is high. Also, the linear classifier even outperforms our method in some case. The result show that original SAM is sensitive to high-frequency perturbations (i.e edges or noise in ultrasound).}
    \label{fig:ultrasound_example}
\end{figure}

We also tested our method on datasets of other modalities.
For example, we tested it on an ultrasound dataset, Pubic Symphysis-Fetal Head Segmentation and Angle of Progression~\cite{jieyun_bai_2023_7851339}, which includes lots of high-frequency features. We test head segmentation using this dataset, see Table~\ref{ultrasound} 
We found that the SAM decoder will lead to degrading performance.  To keep the fairness for both of our methods and other fine-tuning methods, we use k=20 for testing. The result in Table~\ref{ultrasound} shows that our method has a lower performance compared to MedSAM and SAMed. Surprisingly, in the examples in Fig.~\ref{fig:ultrasound_example}, we found that the performance is better when just using the linear classifier and then upscaling. The reason is that the decoder of SAM does not have the capability to predict the accurate mask from the interference of high-frequency features. 
This reflects that although the size and position of the instance are important, the classifier needs to know the basic knowledge of the modality. To solve it, one may combine our methods together with other fine-tuned decoder.

\begin{table}
\caption{The results of our method on the ultrasound dataset: Symphysis-Fetel, in a 20-shot setting. "Linear" means the coarse mask generated by the linear pixel-wise classifier, "point+box" means both of the self-generated point and box are used for the final output.}\label{ultrasound}
\begin{center}
\begin{tabular}{|c|>{\centering\arraybackslash}m{2cm}|>{\centering\arraybackslash}m{2cm}|>{\centering\arraybackslash}m{2cm}|>{\centering\arraybackslash}m{2cm}|}
\hline
\multirow{2}{*}{Models} & \multicolumn{2}{|c|}{Symphysis-Fetal} \\
\cline{2-3}
& Dice & IoU  \\
\hline
Ours(linear) & 66.47 & 53.32 \\
\hline
Ours(point+box) & 69.67 & 55.94 \\

\hline
MedSAM~\cite{ma2023segment} & 73.78 & 61.22  \\
\hline
\end{tabular}
\end{center}
\end{table}
\subsection{Ablation Study Table}
\begin{table}[H]
\caption{Results of the ablation study on Kvasir-SEG. "Linear" means the coarse mask from the linear pixel-wise classifier; "point" means only using the point as prompt; "box" represents only using the bounding box as prompt;"point+box" means using both of bounding box and point as prompt. The scores here are dice scores.}\label{tab4}
\begin{center}
\begin{tabular}{|c|>{\centering\arraybackslash}m{2cm}|>{\centering\arraybackslash}m{2cm}|>{\centering\arraybackslash}m{2cm}|>{\centering\arraybackslash}m{2cm}|}
\hline
\multicolumn{5}{|c|}{Kvasir-SEG} \\
\hline
 & 10 shots & 20 shots & 40 shots & full-data\\
\hline
Ours(linear) & 50.42 & 49.46 & 52.91 & 54.32 \\
\hline
Ours(point) & 58.50 & 62.51 & 63.24 & 63.51 \\
\hline
Ours(box) & 46.55 & 51.95 & 54.91 & 58.12 \\
\hline
Ours(point+box) & 60.03 & 62.78 & 65.34 & 67.08 \\
\hline
\end{tabular}
\end{center}
\end{table}
\end{document}